\documentclass{article}

\usepackage{PRIMEarxiv}

\usepackage[utf8]{inputenc} %
\usepackage[T1]{fontenc}    %
\usepackage{hyperref}       %
\usepackage{url}            %
\usepackage{booktabs}       %
\usepackage{multirow}       
\usepackage{amsfonts}       %
\usepackage{nicefrac}       %
\usepackage{microtype}      %
\usepackage{lipsum}
\usepackage{fancyhdr}       %
\usepackage{graphicx}       %
\usepackage{prettyref}
\usepackage[numbers]{natbib}
\usepackage{amssymb}
\graphicspath{{media/}}     %

\newcommand\megaurl{\url{https://github.com/pmineiro/musique-off}}

\title{Online Joint Fine-tuning of Multi-Agent Flows
}

\author{
  Paul Mineiro \\
  Microsoft Research \\
  \texttt{pmineiro@microsoft.com} \\
}

\begin{document}
\maketitle

\begin{abstract}
A Flow is a collection of component models (``Agents'') which constructs the solution to a complex problem via iterative communication.  Flows have emerged as state of the art architectures for code generation~\cite{karpathytweetflow}, and are the raison d'etre for frameworks like Autogen~\cite{autogen}.  However, flows are currently constructed via a combination of manual prompt engineering and stagewise supervised learning techniques; the latter is limited to acyclic flows with granular node supervision.  In this writeup I describe a procedure for online joint fine-tuning of an entire flow inspired by the Learning to Search framework~\cite{daume2009search}.  The approach leverages simulator access to reduce preferences over entire episodes to preferences over individual node outputs; when the components are language models the latter is a well-studied problem~\cite{rafailov2024direct,rosset2024direct,gorbatovski2024learn}.  The approach is applicable to reward-free settings (e.g., text feedback) if an episode evaluator model is available.  I apply to the multi-hop QA dataset Musique~\cite{trivedi2021musique} achieving a state-of-the-art result.
\end{abstract}

\section{Introduction}
Flows are an emerging state of the art architecture for leveraging foundation models.  I speculate there are multiple reasons for this.  First, iterative computation is capable of correcting errors in ballistic generation~\cite{awesomeselfreflection} and consequently is a strong alternative to Chain-Of-Thought~\cite{wei2022chain}.  Second, specialization of components eases the burden of prompt engineering and allows humans to inject prior information via intuitive task decomposition.  Finally the use of specialized components facilitates adding functionality to a system over time: Musique provides an explicit example of this, as discussed in \prettyref{sec:discussion}.

In this writeup I provide one approach for flow self-improvement based upon access to an environment simulator, i.e., the ability to reset state to an intermediate point in the computation and evaluate alternative outputs.\footnote{A simulator can typically be constructed from a machine-learning dataset, as in the Musique experiments.}  This procedure might be invoked after hand-tuning of a specific flow yields diminishing returns; alternatively, more complex flows can be initially considered as the burden of hand-specification is reduced.  The key idea is to reduce episode preferences to individual node output preferences by simulating the impact of changing the output of a single node.  Individual node output preferences can be optimized using parametric preference learning if all nodes are language models~\cite{rafailov2024direct,rosset2024direct,gorbatovski2024learn}; other non-language-model components can also utilize pairwise preference data, e.g., ranking systems~\cite{furnkranz2003pairwise}.

Fortunately, the approach does not require rewards per se, only the ability to express a preference over entire episodes.  When rewards are available, this is a natural way to express a preference; but if they are not available (e.g., the environment only provides text feedback) than a (large foundation) model can be used to induce a preference.  The Musique example described in \prettyref{sec:musique} is a traditional dataset with a (multiobjective) reward function so there is not an explicit example of this herein, but the exposition will proceed generally.

The reader should consider this document a white paper describing preliminary results, rather than a mature scientific publication.  The primary purpose of this document is to facilitate replication of the state-of-the-art Musique result.  Although the approach is plausibly broadly applicable, systematic evaluation on a wider range of problems is necessary before drawing any strong conclusions.

\begin{figure}[!t]
\vskip 0pt
{
    \centering
    \includegraphics[width=.99\textwidth]{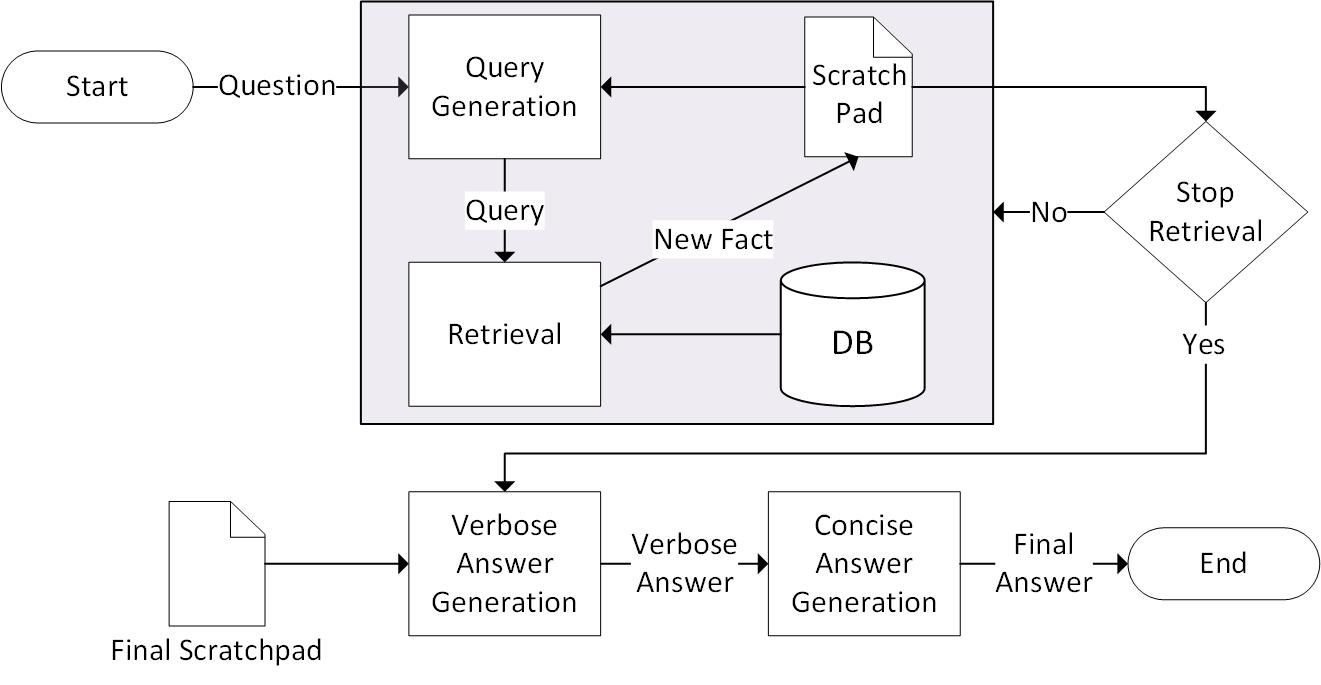}
    \caption{Actual flow utilized for Musique-Answerable.  Key realistic aspects are: dependent generation with loops (the entire retrieval block is looped multiple times); dynamic conversation state (``scratchpad''); learned conditional branching (``stop retrieval''); and lack of granular supervision.}
    \label{fig:musique}
}
\end{figure}

\section{The Approach}

\subsection{The Challenges}

\prettyref{fig:musique} diagrams the actual flow utilized for the submission to the easier variant of the Musique leaderboard (``Musique Answerable'').  A more detailed description is provided in \prettyref{sec:musique}.  Here I highlight several realistic properties which create difficulties for flow learning.  The first is looping, which makes it awkward to apply stagewise learning procedures: when a node is updated, this changes the distribution of inputs to that node.  The second is dynamic conversation state and conditional branching, which cascades to different downstream inputs and outputs encountered by other nodes.  Finally the lack of granular supervision implies only the result of the joint sequence of decisions of all nodes can be evaluated.  These challenges are not unique to flow learning and are idiosyncratic instantiations of the general difficulties of reinforcement learning.

Interestingly, many benchmark datasets include additional metadata to facilitate granular supervision, and Musique is no exception: it includes {\tt question decomposition} metadata that could be used to provide granular supervision to the query generator node in \prettyref{fig:musique}.  Naive use of such information can exhibit the compounding errors problem associated with behaviour cloning~\cite{wenimitation}.  Furthermore in practice such granular supervision either does not exist or is tedious to provide, so it's better to have a method that does not require it.  There is no free lunch, however: the procedure pays both a computational price (evaluating rollouts during learning) and a statistical price (reliable credit assignment over complex flows can require prohibitively large amounts of exploration).  Mitigating the latter issue requires initialization in a rewarding region of policy space: see paragraph Prompts in \prettyref{sec:musique} below.

\begin{figure}[!t]
\vskip 0pt
{
    \centering
    \includegraphics[width=.99\textwidth]{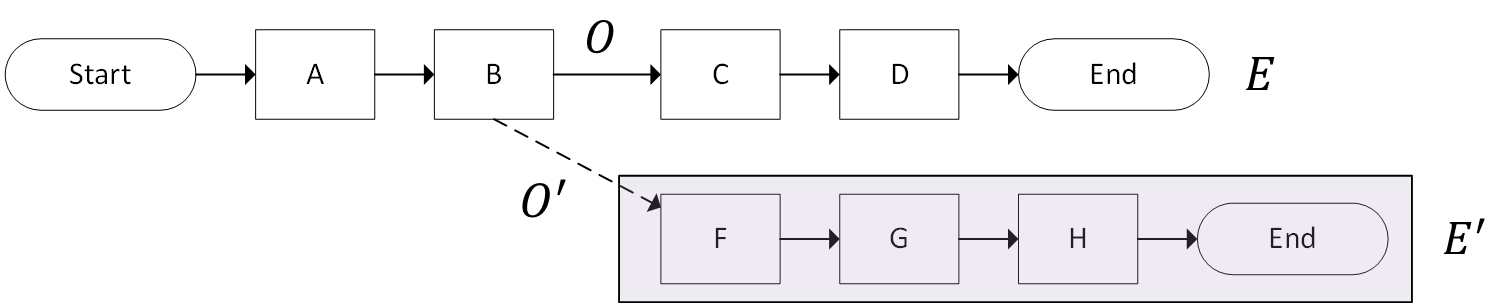}
    \caption{Visualization of one-step deviations for inducing individual node output preferences $(O' \succ O)$ from complete episode preferences $(E' \succ E)$.  At each point in the realized unrolled computation graph, an alternative output is produced and then the episode is evaluated until termination.  Note this can result in a different sequence of nodes being called due to conditional execution.  Furthermore, although all the node identifiers in this picture are unique, in practice the same node appears at multiple point in the unrolled computation graph due to loops.  Finally, rollouts do not themselves generate rollouts, so the computational overhead of learning is quadratic in the length of the unrolled computation graph (a linear number of rollouts, each doing a linear amount of additional inference).}
    \label{fig:rollouts}
}
\end{figure}

\subsection{Addressing the Challenges}

The core idea of the approach is inspired Learning To Search~\cite{daume2009search}, and borrows two key ideas.  The first and less important idea is ``optimizing for greedy inference.''  In other words, the goal of learning is to create a flow in which each node produces a single output at inference time, but which nonetheless produces a desired episode outcome.  This superficially constrasts with approaches that explicitly use planning or backtracking search at inference time, e.g., MCTS-LLM~\cite{zhao2024large}.  However this is not a hard distinction, as (re)planning nodes and conditional execution can be incorporated into the flow; nonetheless envisioning greedy inference makes the use of one-step deviations more intuitive.

The second borrowed idea is the use of one-step deviations to solve the credit assignment problem.  As indicated in \prettyref{fig:rollouts}, at each point in the unrolled realized computation graph, an alternative output is produced, and then rolled out until the end of the episode.  This reduces preferences over entire episode outcomes into preferences over individual node outputs.  For large language models, optimizing preferences over individual outputs is an area of active research with frequent algorithmic innovations, e.g., \citet{rafailov2024direct}, \citet{rosset2024direct}, and \citet{gorbatovski2024learn}.

\paragraph{How to Generate Rollouts?}  I've had good luck on other problems using a single alternative rollout generated heuristically using the {\tt diverse\_beam} generation~\cite{vijayakumar2016diverse} built into Huggingface Transformers~\cite{transformers}.  However for best performance on Musique I had to increase the number of rollouts, and this was done empirically by monitoring progressive validation loss~\cite{blum1999beating} during training.  A better exploration strategy could yield benefits.

\subsection{Technical Errata}

Because of the previously mentioned distributional shifts associated with node updates, I use an online variants of DPO preference learning, i.e., nodes are immediately updated after a (mini-batch) of episodes is evaluated.  I find the (parameter free) Pytorch Coin Betting Optimizer~\cite{coinbettinggithub} is adept at navigating the internal nonstationarity induced by the training procedure, although conceivably other optimizers might work with sufficient hyperparameter tweaking.

If direct final supervision is available, such as in Musique, then for (only) the terminal node of the computation graph it is possible and better to use supervised learning rather than preference learning.  Specifically this is used for the {\tt Concise Answer Generation} node in \prettyref{fig:musique}, as the dataset includes a gold-label final answer.  More generally, rollouts can be terminated early if episode preferences can be induced from partial trajectories: this is the case in Musique due to the multiobjective nature of the reward function, as described in \prettyref{sec:musique}.

In practice each node in the computation graph is a parameter efficient fine-tuned adaptor~\cite{peft} layered onto to a common instruction-tuned foundation model.  I only experimented with Lora~\cite{hu2021lora} as I found it sufficient for good performance.

The complete implementation for Musique can be found at \megaurl.

\section{Musique Results}
\label{sec:musique}

The Musique dataset~\cite{trivedi2021musique} is grounded question-answering benchmark which assess both the quality of the answer and accuracy identifying the (lack of) supporting material.  It comes in two variants: the ``Musique-Answerable`` version in which every question can be answered using the documents associated with the example; and the ``Musique-Full`` version, in which it is also necessary to determine if the documents associated with the example are sufficient to answer the question.  The criterion for the dataset is multi-objective: for answerable questions, the final answer is scored based upon largest ``answer f1'' overlap\footnote{The answer is tokenized and the set of tokens is compared to the gold set of tokens to compute precision and recall.} to the (possibly multiple) gold-label answers; furthermore, the set of retrieved documents is scored for overlap with the gold standard set (``support f1'').  Finally, for the unanswerable variant, the accuracy of the sufficiency prediction is also incorporated.  Multi-objective reinforcement learning is in general quite complicated, as it is only possible to identify policies that are not Pareto dominated~\cite{hayes2022practical}.  For Musique we sidestep this difficulty by rewarding the retrieval step on support f1, the answer step on answer f1, and the sufficiency step on sufficiency accuracy.  While this ignores the interactions between these components of the reward, it turns out excellent support f1 and sufficiency accuracy are possible, and therefore this works well.  This also allows for early termination of rollouts which improves the computational efficiency.

\begin{figure}[!t]
\vskip 0pt
{
    \centering
    \includegraphics[width=.99\textwidth]{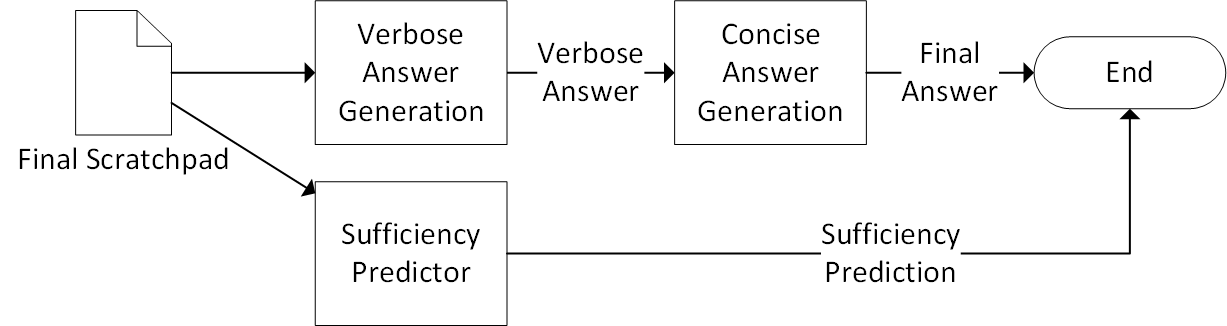}
    \caption{For the Full version of the Musique dataset, an additional node is introduced into the flow to predict sufficiency.  The flow is otherwise identical.}
    \label{fig:answerablevsfull}
}
\end{figure}

Unfortunately standard distributed training frameworks prescribe a (simplistic) control flow which is not convenient when worker threads are evaluating the dynamic call graphs of realistic flows; the solution is an indirection layer between episode evaluation and model invocation which makes the training nonstandard and hence difficult to read.  Fortunately the function \href{https://github.com/pmineiro/musique-off/blob/8415d4c5f7b72772344bb410cc6e3acdc28724cf/answerable/qdecompdyn.py#L3}{\tt qdecompdyn.py:evaluate\_conversation()} can be understood in isolation as the realization of \prettyref{fig:musique}.  The design includes the following components:
\begin{itemize}
\item \textbf{Query Generation}: given the original question and the current scratchpad of documents, generate a query for the retrieval system to return a new document. \\
\item \textbf{Retrieval}: given a query and a set of candidate documents, rank them by their utility in answering the query, and return the best document. \\
\item \textbf{Stop Retrieval}: given the original query and the current scratchpad of documents, decide whether additional retrieval is necessary. \\
\item \textbf{Verbose Answer Generation}: given the original query and the current scratchpad of documents, generate a verbose answer. \\
\item \textbf{Concise Answer Generation}: given the original query and the verbose answer, generate a concise answer. \\
\item \textbf{Sufficiency} (``Musique-Full'' only): given the original query and the final scratchpad of documents, decide if the query could be answered given the database.  The modification is diagrammed in \prettyref{fig:answerablevsfull}.
\end{itemize}
The two stage answer design is beneficial because the Musique objective function penalizes loquaciousness, and initial experimentation indicated it outperformed direct generation of the concise answer.  In a realistic deployment, I believe users would prefer the verbose answer.

\paragraph{Prompts} \label{par:prompts} In practice, the sample efficiency of reinforcement learning is related to the density of rewards experienced under random rollouts~\cite{laidlaw2023bridging}.  This procedure also appears subject to this, as it is empirically helpful to choose initial prompts for the components that initialize the flow in a rewarding region.  The specific prompts used are found in \href{https://github.com/pmineiro/musique-off/blob/8415d4c5f7b72772344bb410cc6e3acdc28724cf/answerable/Prompts.py#L1}{\tt Prompts.py}.  I did not obsess over them, but it was important to choose reasonable initial prompts so that components had outputs with some correspondence to their desired roles in the computation.  The same prompts were used for all language models without additional customization under the assumption that fine-tuning would mitigate model differences.  

For both Musique-Answerable and Musique-Full, the online fine-tuned flow achieves a new state-of-the-art result at the time of this writing.  As indicated in \prettyref{tab:finetune}, the untrained performance of this flow is non-zero, suggesting that the prompts initialize the flow near a highly rewarding region; but there are  substantial benefits from fine-tuning.
\begin{table}[t]
\centering
\begin{tabular}{lcccccc}\toprule
& \multicolumn{2}{c}{Progressive Validation} & \multicolumn{2}{c}{Progressive Validation} \\
Flow & \multicolumn{2}{c}{On Training Set} & \multicolumn{2}{c}{On Dev Set} & \multicolumn{2}{c}{Test Set} \\
\cmidrule(lr){2-3} \cmidrule(lr){4-5} \cmidrule{6-7}
& AnsF1 & SuppF1 & AnsF1 & SuppF1 & AnsF1 & SuppF1 \\ \midrule
Untrained & 0.085 & 0.492 & 0.074 & 0.437 & - & - \\
+Train on Train & 0.729 & 0.904 & 0.629 & 0.887 & - & - \\
+Enforce Grounded & - & - & 0.650 & 0.887 & - & - \\
+Train on Dev & - & - & 0.761 & 0.92 & 0.693 & 0.92 \\ \bottomrule
\end{tabular}
\vskip 12pt
\caption{Assorted treatment effects for Musique Answerable.  ``$-$'' indicates ``not measured''.  \emph{Fine-tuning improves performance}:  Although the untrained performance is quite poor, it is empirically sufficient to initialize the flow such that learning can succeed. \emph{Enforcing grounded answers improves AnsF1}: Post-processing the output of the flow to correspond to a span in a selected document is beneficial.  \emph{Training on the Dev set is internally consistent}: Progressive validation loss on the dev set improves, suggesting both additional training is not harmful and the dev set is self-redundant.}
\label{tab:finetune}
\end{table}

\subsection{Hyperparameter Selection}

The use of Coin Betting eliminates hyperparameters associated with optimization per se, but there are additional hyperparameters to be tuned.  Musique is a large dataset and good performance is possible after a single pass over the training set\footnote{Furthermore, in early experimentation, a second pass over the training set had no impact on performance on the validation set; therefore I never did more than one training pass.}, so I used progressive validation loss to make decisions when possible, as this allowed for early termination of experiments associated with bad ideas: see ``Subsampling the Training Set'' in \prettyref{subsec:musiquespecific} below.

Using periodic soft ref updating~\cite{gorbatovski2024learn} with a larger DPO regularization parameter ($\beta = 1$) was better than using a fixed reference policy and a lower DPO regularization parameter ($\beta = 0.1$).  I gated the soft ref update on progressive validation improvement in answer f1, because this seemed like a good idea; I have not tested whether unconditional soft ref updating yields better results.

I tried models {\tt mistralai/Mistral-7B-Instruct-v0.2}~\citep{jiang2023mistral}, {\tt meta-llama/Meta-Llama-3-8B-Instruct}~\citep{llama}, {\tt Qwen/Qwen1.5-32B-Chat}~\citep{qwen}, and {\tt microsoft/Phi-3-medium-128k-instruct}~\citep{phi}; the latter had the best performance. The best rank of the Lora adapter for different models was determined by trying all values $\{ 2, 4, 8, 16, 32 \}$; $8$ gave the best results for Phi.  I did limited experimentation with models as large as {\tt meta-llama/Meta-Llama-3-70B-Instruct} but such experiments ran slowly and did not exhibit substantial lift in progressive validation loss during the initial part of training, so I early terminated those experiments and focused on the smaller models.\footnote{In my experience with fine-tuning flows for other problems, the performance advantage of large models can be substantial before training but is typically eliminated by fine-tuning.}

\subsection{Musique Specific Techniques}
\label{subsec:musiquespecific}

\paragraph{Subsampling the Training Set}  The training set of Musique contains a different distribution of ``hops'' (number of reasoning steps required to answer the question) than the dev and test sets, see Table 2 of \citep{trivedi2021musique}.  Subsampling the training set to match these moments reduces training time by two-thirds and brings progressive validation loss on the training set more into correspondence with evaluation on the dev set.  With subsampling, it takes an 8xH100 machine about a day to consume the answerable training set.

\paragraph{Enforcing Grounded Answers} Musique is intended for grounded question-answering and not regurgitation of (potentially inaccurate) pre-trained knowledge.  Therefore I postprocess the output of the flow to correspond to a span in one of the selected documents.  This has no effect on support f1 or sufficiency metrics, but yields an improvement in answer f1 on the validation set as indicated in \prettyref{tab:finetune}.

\paragraph{Train on Dev} With online learning, one can monitor progressive validation loss; consequently there is no need to have a distinct data split to prevent overfitting.  Because data splits are inefficient, I chose to train on the dev set before submission.  As indicated in \prettyref{tab:finetune}, this yields improvement on the dev set (assessed progressively): this suggests both that additional training was not harmful and the dev set is self-redundant.  Without abusing the submission oracle, I have no way to determine whether this additional training step is helpful or not, but I assumed it was and used it in the final submission. This assumption may be incorrect because the test set is designed to be out-of-distribution with the training and dev sets.

\paragraph{Pairwise Inference} On the Full version of the dataset, the questions come in pairs, exactly one of which is answerable and the other is not.  Enforcing this constraint (by sorting on the output of the sufficiency node in the flow) yields substantial improvement in sufficiency accuracy, which improves the combined f1+sufficiency leaderboard metrics.  This is quantified in \prettyref{tab:finetunefull}.

\paragraph{Miscellaneous} The maximum number of retrievals was capped at 4, because this was the minimum value which achieved near perfect support f1, and larger values ran more slowly.  All base models were 4-bit double quantized using {\tt bitsandbytes}~\citep{dettmers2023case,dettmers2023qlora}.

\begin{table}[t]
\centering
\begin{tabular}{lcccccccc}\toprule
& \multicolumn{3}{c}{Progressive Validation} & \multicolumn{3}{c}{Progressive Validation} \\
Flow & \multicolumn{3}{c}{On Training Set} & \multicolumn{3}{c}{On Dev Set} & \multicolumn{2}{c}{Test Set} \\
\cmidrule(lr){2-4} \cmidrule(lr){5-7} \cmidrule{8-9}
& AnsF1 & SuppF1 & SuffAcc & AnsF1 & SuppF1 & SuffAcc & Ans+SuffF1 & Supp+SuffF1 \\ \midrule
Untrained & 0.073 & 0.445 & 0.627 & 0.074 & 0.448 & 0.623 & - & - \\
+Train on Train & 0.689 & 0.887 & 0.86 & 0.579 & 0.883 & 0.874 & - & - \\
+Enforce Grounded & - & - & - & 0.604 & 0.883 & 0.874 & - & - \\
+Pairwise Inference & - & - & - & 0.604 & 0.883 & 0.945 & - & - \\
+Train on Dev & - & - & - & 0.740 & 0.921 & 0.957 & 0.672 & 0.887 \\ \bottomrule
\end{tabular}
\vskip 12pt
\caption{Assorted treatment effects for Musique Full.  ``$-$'' indicates ``not measured''.  ``Ans+SuffF1'' (``Supp+SuffF1'') is computed by the leaderboard evaluation and combines AnsF1 (SuppF1, resp.) with Sufficiency Accuracy.  \emph{Pairwise inference improves sufficiency accuracy}: Enforcing the constraint that exactly one of each paired example is answerable is beneficial.  \emph{Other treatment effects}: See the caption of \prettyref{tab:finetune} for explanation.}
\label{tab:finetunefull}
\end{table}

\section{Discussion}
\label{sec:discussion}

This work was inspired by several observations: 1) flows are SOTA architectures for multiple benchmark tasks of interest; 2) flows are currently mostly hand-engineered; 3) flows are typically short, i.e., the unrolled realized computation graph of a typical flow has 10 or less nodes in it; and 4) pre-trained and instruction-tuned foundation models induce reasonable prior distributions over actions using ``obvious'' prompts for initialization.  This combination of factors motivates the application of model-free short-horizon online reinforcement learning techniques. If any of these factors are relaxed, we can expect the approach described here to exhibit multiple failure modes.  As an extreme example, using a tabula rasa transformer model would presumably result in most rollouts yielding no episode preferences; and even if training were run long enough to generate successful rollouts, I suspect the procedure would ultimately overfit the training set and not generalize well.  A more likely failure mode to be encountered in practice is a problem which is well-covered by pre-training (e.g., an automated software engineering flow) but requires a long execution chain to succeed (e.g., circa 50 node invocations); in this case, rollouts may still mostly fail to find improvements without a high-quality initialization of the flow; and the one-step-deviation credit assignment strategy (which corresponds to a local policy improvement guarantee) may not be powerful enough to converge to a strong result.

The approach describe here requires simulator access, i.e., the ability to reset state to an intermediate point in the computation and evaluate alternative outputs.  This is applicable both to offline tuning and some online production scenarios (e.g., a software development assistant could plausibly reset state and evaluate an alternative rollout); however, for self-improvement from production deployment, a single-sample approach is often required.  For example, a website powered by a flow cannot ask the user to start over halfway through their session so that the algorithm can try something different and see if it works better.
I'm currently investigating single-sample approaches.

There is a relative lack of submission activity for Musique-Full compared to Musique-Answerable.  Why don't researchers take their solutions for Musique-Answerable and also run them on Musique-Full? This highlights an important benefit of flow based architectures: if you want to add capabilities to an existing system over time, having specialized components is beneficial. Here, the difference between the two is simply adding a single node to the flow as in \prettyref{fig:answerablevsfull}.

\bibliographystyle{unsrtnat}
\bibliography{references}  

\begin{thebibliography}{26}
\providecommand{\natexlab}[1]{#1}
\providecommand{\url}[1]{\texttt{#1}}
\expandafter\ifx\csname urlstyle\endcsname\relax
  \providecommand{\doi}[1]{doi: #1}\else
  \providecommand{\doi}{doi: \begingroup \urlstyle{rm}\Url}\fi

\bibitem[kar()]{karpathytweetflow}
{Prompt engineering (or rather "Flow engineering") intensifies for code
  generation}.
\newblock
  \url{https://twitter.com/karpathy/status/1748043513156272416?lang=en}.
\newblock Accessed: 2024-05-14.

\bibitem[aut()]{autogen}
{AutoGen: Enable Next-Gen Large Language Model Applications}.
\newblock \url{https://microsoft.github.io/autogen/}.
\newblock Accessed: 2024-05-14.

\bibitem[Daum{\'e} et~al.(2009)Daum{\'e}, Langford, and Marcu]{daume2009search}
Hal Daum{\'e}, John Langford, and Daniel Marcu.
\newblock Search-based structured prediction.
\newblock \emph{Machine learning}, 75:\penalty0 297--325, 2009.

\bibitem[Rafailov et~al.(2024)Rafailov, Sharma, Mitchell, Manning, Ermon, and
  Finn]{rafailov2024direct}
Rafael Rafailov, Archit Sharma, Eric Mitchell, Christopher~D Manning, Stefano
  Ermon, and Chelsea Finn.
\newblock Direct preference optimization: Your language model is secretly a
  reward model.
\newblock \emph{Advances in Neural Information Processing Systems}, 36, 2024.

\bibitem[Rosset et~al.(2024)Rosset, Cheng, Mitra, Santacroce, Awadallah, and
  Xie]{rosset2024direct}
Corby Rosset, Ching-An Cheng, Arindam Mitra, Michael Santacroce, Ahmed
  Awadallah, and Tengyang Xie.
\newblock Direct nash optimization: Teaching language models to self-improve
  with general preferences.
\newblock \emph{arXiv preprint arXiv:2404.03715}, 2024.

\bibitem[Gorbatovski et~al.(2024)Gorbatovski, Shaposhnikov, Malakhov,
  Surnachev, Aksenov, Maksimov, Balagansky, and Gavrilov]{gorbatovski2024learn}
Alexey Gorbatovski, Boris Shaposhnikov, Alexey Malakhov, Nikita Surnachev,
  Yaroslav Aksenov, Ian Maksimov, Nikita Balagansky, and Daniil Gavrilov.
\newblock Learn your reference model for real good alignment.
\newblock \emph{arXiv preprint arXiv:2404.09656}, 2024.

\bibitem[Trivedi et~al.(2022)Trivedi, Balasubramanian, Khot, and
  Sabharwal]{trivedi2021musique}
Harsh Trivedi, Niranjan Balasubramanian, Tushar Khot, and Ashish Sabharwal.
\newblock {M}u{S}i{Q}ue: Multihop questions via single-hop question
  composition.
\newblock \emph{Transactions of the Association for Computational Linguistics},
  2022.

\bibitem[awe()]{awesomeselfreflection}
{Awesome LLM Self-Reflection}.
\newblock \url{https://github.com/rxlqn/awesome-llm-self-reflection}.
\newblock Accessed: 2024-05-14.

\bibitem[Wei et~al.(2022)Wei, Wang, Schuurmans, Bosma, Xia, Chi, Le, Zhou,
  et~al.]{wei2022chain}
Jason Wei, Xuezhi Wang, Dale Schuurmans, Maarten Bosma, Fei Xia, Ed~Chi, Quoc~V
  Le, Denny Zhou, et~al.
\newblock Chain-of-thought prompting elicits reasoning in large language
  models.
\newblock \emph{Advances in neural information processing systems},
  35:\penalty0 24824--24837, 2022.

\bibitem[F{\"u}rnkranz and H{\"u}llermeier(2003)]{furnkranz2003pairwise}
Johannes F{\"u}rnkranz and Eyke H{\"u}llermeier.
\newblock Pairwise preference learning and ranking.
\newblock In \emph{European conference on machine learning}, pages 145--156.
  Springer, 2003.

\bibitem[wen()]{wenimitation}
{Wen Sun: Imitation Learning Lecture}.
\newblock
  \url{https://wensun.github.io/CS4789_data/Imitation_Learning_April_8_annotated.pdf}.
\newblock Accessed: 2024-05-14.

\bibitem[Zhao et~al.(2024)Zhao, Lee, and Hsu]{zhao2024large}
Zirui Zhao, Wee~Sun Lee, and David Hsu.
\newblock Large language models as commonsense knowledge for large-scale task
  planning.
\newblock \emph{Advances in Neural Information Processing Systems}, 36, 2024.

\bibitem[Vijayakumar et~al.(2016)Vijayakumar, Cogswell, Selvaraju, Sun, Lee,
  Crandall, and Batra]{vijayakumar2016diverse}
Ashwin~K Vijayakumar, Michael Cogswell, Ramprasath~R Selvaraju, Qing Sun,
  Stefan Lee, David Crandall, and Dhruv Batra.
\newblock Diverse beam search: Decoding diverse solutions from neural sequence
  models.
\newblock \emph{arXiv preprint arXiv:1610.02424}, 2016.

\bibitem[tra()]{transformers}
{Transformers}.
\newblock \url{https://github.com/huggingface/transformers}.
\newblock Accessed: 2024-05-14.

\bibitem[Blum et~al.(1999)Blum, Kalai, and Langford]{blum1999beating}
Avrim Blum, Adam Kalai, and John Langford.
\newblock Beating the hold-out: Bounds for k-fold and progressive
  cross-validation.
\newblock In \emph{Proceedings of the twelfth annual conference on
  Computational learning theory}, pages 203--208, 1999.

\bibitem[coi()]{coinbettinggithub}
{Parameter-Free Optimizers for PyTorch}.
\newblock \url{https://github.com/bremen79/parameterfree}.
\newblock Accessed: 2024-05-14.

\bibitem[pef()]{peft}
{State-of-the-art Parameter-Efficient Fine-Tuning (PEFT) methods}.
\newblock \url{https://github.com/huggingface/peft}.
\newblock Accessed: 2024-05-14.

\bibitem[Hu et~al.(2021)Hu, Shen, Wallis, Allen-Zhu, Li, Wang, Wang, and
  Chen]{hu2021lora}
Edward~J Hu, Yelong Shen, Phillip Wallis, Zeyuan Allen-Zhu, Yuanzhi Li, Shean
  Wang, Lu~Wang, and Weizhu Chen.
\newblock Lora: Low-rank adaptation of large language models.
\newblock \emph{arXiv preprint arXiv:2106.09685}, 2021.

\bibitem[Hayes et~al.(2022)Hayes, R{\u{a}}dulescu, Bargiacchi,
  K{\"a}llstr{\"o}m, Macfarlane, Reymond, Verstraeten, Zintgraf, Dazeley,
  Heintz, et~al.]{hayes2022practical}
Conor~F Hayes, Roxana R{\u{a}}dulescu, Eugenio Bargiacchi, Johan
  K{\"a}llstr{\"o}m, Matthew Macfarlane, Mathieu Reymond, Timothy Verstraeten,
  Luisa~M Zintgraf, Richard Dazeley, Fredrik Heintz, et~al.
\newblock A practical guide to multi-objective reinforcement learning and
  planning.
\newblock \emph{Autonomous Agents and Multi-Agent Systems}, 36\penalty0
  (1):\penalty0 26, 2022.

\bibitem[Laidlaw et~al.(2023)Laidlaw, Russell, and Dragan]{laidlaw2023bridging}
Cassidy Laidlaw, Stuart~J Russell, and Anca Dragan.
\newblock Bridging rl theory and practice with the effective horizon.
\newblock \emph{Advances in Neural Information Processing Systems},
  36:\penalty0 58953--59007, 2023.

\bibitem[Jiang et~al.(2023)Jiang, Sablayrolles, Mensch, Bamford, Chaplot,
  Casas, Bressand, Lengyel, Lample, Saulnier, et~al.]{jiang2023mistral}
Albert~Q Jiang, Alexandre Sablayrolles, Arthur Mensch, Chris Bamford,
  Devendra~Singh Chaplot, Diego de~las Casas, Florian Bressand, Gianna Lengyel,
  Guillaume Lample, Lucile Saulnier, et~al.
\newblock Mistral 7b.
\newblock \emph{arXiv preprint arXiv:2310.06825}, 2023.

\bibitem[lla()]{llama}
{Introducing Meta Llama 3: The most capable openly available LLM to date}.
\newblock \url{https://ai.meta.com/blog/meta-llama-3/}.
\newblock Accessed: 2024-05-14.

\bibitem[qwe()]{qwen}
{Introducing Qwen1.5}.
\newblock \url{https://qwenlm.github.io/blog/qwen1.5/}.
\newblock Accessed: 2024-05-14.

\bibitem[phi()]{phi}
{Introducing Phi-3: Redefining what’s possible with SLMs}.
\newblock
  \url{https://azure.microsoft.com/en-us/blog/introducing-phi-3-redefining-whats-possible-with-slms/}.
\newblock Accessed: 2024-05-14.

\bibitem[Dettmers and Zettlemoyer(2023)]{dettmers2023case}
Tim Dettmers and Luke Zettlemoyer.
\newblock The case for 4-bit precision: k-bit inference scaling laws.
\newblock In \emph{International Conference on Machine Learning}, pages
  7750--7774. PMLR, 2023.

\bibitem[Dettmers et~al.(2023)Dettmers, Pagnoni, Holtzman, and
  Zettlemoyer]{dettmers2023qlora}
Tim Dettmers, Artidoro Pagnoni, Ari Holtzman, and Luke Zettlemoyer.
\newblock Qlora: Efficient finetuning of quantized llms.
\newblock \emph{arXiv preprint arXiv:2305.14314}, 2023.

\end{thebibliography}
\end{document}